\documentclass[10pt,twocolumn,letterpaper]{article}

\usepackage[pagenumbers]{iccv} 
\usepackage{multirow}
\definecolor{iccvblue}{rgb}{0.21,0.49,0.74}
\usepackage[pagebackref,breaklinks,colorlinks,allcolors=iccvblue]{hyperref}

\def\paperID{6104} 
\def\confName{ICCV}
\def\confYear{2025}

\title{Multi-identity Human Image Animation with Structural Video Diffusion}

\author{  Zhenzhi Wang$^1$, Yixuan Li$^1$, Yanhong Zeng$^2$, Yuwei Guo$^1$, Dahua Lin$^{1,2}$, Tianfan Xue$^{1,2}$, Bo Dai$^3$ \\
  $^1$The Chinese University of Hong Kong, $^2$Shanghai Artificial Intelligence Laboratory, \\ $^3$The University of Hong Kong \\
  \small\texttt{\{wz122, ly122, gy023, dhlin, tfxue\}@ie.cuhk.edu.hk, zengyh1900@gmail.com, bdai@hku.hk}
}
\newcommand{\modelname}{Structural Video Diffusion}
\begin{document}
\maketitle

\begin{figure*}  
    \centering  
    \includegraphics[width=0.95\textwidth]{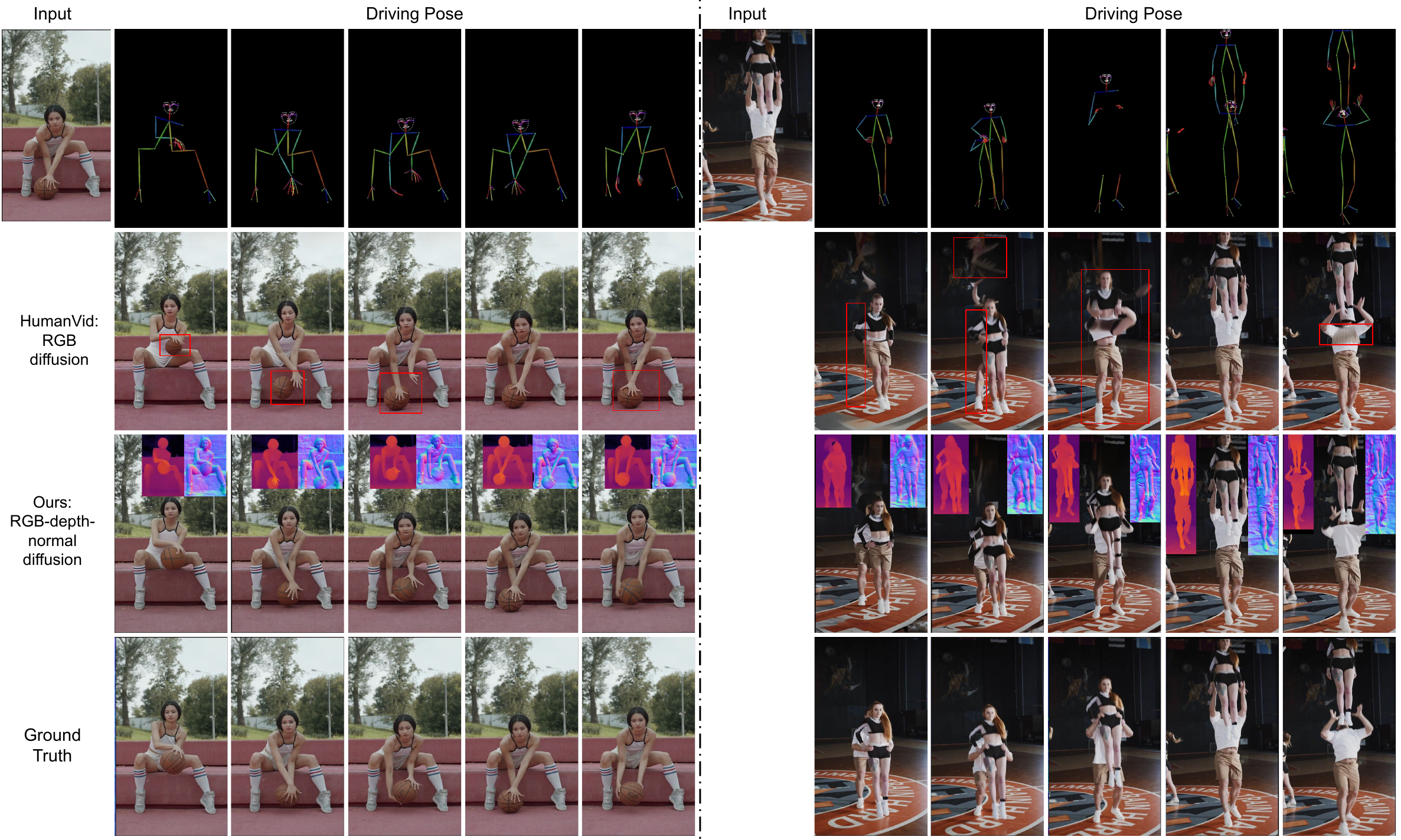}
    \caption{Illustration of multi-identity human image animation in the (left) human-object interaction and (right) multi-human interaction scenarios. Comparing with previous SoTA methods~\cite{wang2024humanvid}, our method shows better quality and pose-following in regions with interactions via joint learning of RGB, depth and surface-normal maps. Since the baseline normal estimation method~\cite{DBLP:conf/eccv/KhirodkarBMZJSAS24} is limited to predicting human normal maps, we restrict our model's supervision to normal maps within human masks. {\bf Best viewed in color and zoom in.}}  
    \label{fig:teaser}  
    \vspace{-1em}
\end{figure*} 

\begin{abstract}
Generating human videos from a single image while ensuring high visual quality and precise control is a challenging task, especially in complex scenarios involving multiple individuals and interactions with objects. Existing methods, while effective for single-human cases, often fail to handle the intricacies of multi-identity interactions because they struggle to associate the correct pairs of human appearance and pose condition and model the distribution of 3D-aware dynamics. To address these limitations, we present \emph{Structural Video Diffusion}, a novel framework designed for generating realistic multi-human videos. Our approach introduces two core innovations: identity-specific embeddings to maintain consistent appearances across individuals and a structural learning mechanism that incorporates depth and surface-normal cues to model human-object interactions. Additionally, we expand existing human video dataset with 25K new videos featuring diverse multi-human and object interaction scenarios, providing a robust foundation for training. Experimental results demonstrate that Structural Video Diffusion achieves superior performance in generating lifelike, coherent videos for multiple subjects with dynamic and rich interactions, advancing the state of human-centric video generation. Code is available at \href{https://github.com/zhenzhiwang/Multi-HumanVid}{here}. 
\end{abstract}    
\section{Introduction}
\label{sec:intro}

Human image animation aims to generate high-fidelity human videos from a single image and a driving sequence of controls, such as poses \cite{hu2023animate,guo2023animatediff,zhu2024champ} or camera viewpoints \cite{wang2024humanvid}. This field has garnered significant attention for its applications in film, gaming, and other creative industries. Despite recent progress, robustly animating humans in complex, real-world scenarios—particularly those involving multiple identities—remains a significant challenge.

While recent video diffusion models have achieved significant advancements in single-person animation \cite{xu2023magicanimate,hu2023animate,zhang2024mimicmotion,yang2024cogvideox}, they falter in multi-identity settings, often producing severe artifacts. These failures highlight two critical challenges: maintaining the appearance consistency of each individual and modeling realistic 3D-aware interactions. In scenes with \textbf{human-human} interactions, such as partner dancing, existing methods struggle to preserve individual appearances and coordinate complex motions. For \textbf{human-object} interactions, objects often appear blurry, float unnaturally, or vanish during movement. These limitations arise from a lack of fine-grained identity control and dedicated mechanisms for modeling 3D spatial relationships, issues we must address to enable more practical applications.

In this paper, we introduce \modelname, a framework for multi-identity human animation that ensures both precise identity control and realistic 3D-aware interactions. Our key insight is that trackable, identity-specific features are essential for appearance consistency, while geometric cues are crucial for modeling spatial relationships. Accordingly, \modelname~introduces two core innovations: (1) ID-Specific Embedding Learning, which uses mask-guided tokens to represent and track each individual, and (2) Latent Structure Learning, which jointly models RGB and geometric information. This dual approach enables \modelname~to maintain consistent appearances during dynamic motion and to accurately handle complex spatial interactions, from partner dancing to object manipulation.

To maintain appearance consistency as individuals move and interact, we introduce learnable ID-specific embeddings tied to segmentation masks. First, we use Segment-Anything V2 (SAM2)~\cite{ravi2024sam} to generate masks for each person. During training, we learn a unique embedding for each identity and use it to populate the corresponding masked region, creating an ID-aware feature map. This mechanism ensures a persistent association between an identity and its appearance, even when subjects swap positions. By representing each person with an independently trackable token derived from a single multi-person reference image, our model preserves identity integrity throughout the generated video. The framework flexibly supports up to N unique identities and can generate videos with any subset of them.

To model complex interactions that go beyond pose guidance, we incorporate geometric cues. Instead of requiring impractical frame-wise depth and surface-normal maps as inputs, we propose a joint learning framework where these maps are predicted alongside the RGB video. This forces the model to become implicitly aware of the underlying 3D structure. For instance, depth maps provide crucial cues for occlusion and relative distances, while surface normals help preserve the shape of objects and clothing during motion. During training, these geometric maps are generated using off-the-shelf estimators~\cite{hu2024depthcrafter,DBLP:conf/eccv/KhirodkarBMZJSAS24} and serve as additional supervision signals. By learning to denoise RGB, depth, and normal information jointly, our model captures the coupled dynamics of appearance and geometry, significantly improving its ability to model realistic human-object interactions, such as a person holding a mug.

To facilitate research on complex multi-person interactions, we introduce the Multi-HumanVid dataset. We extend the recent HumanVid dataset~\cite{wang2024humanvid} with 25,000 new high-quality videos featuring rich human-human and human-object interactions. In addition to the camera parameters~\cite{wang2024tram} and human poses~\cite{yang2023effective} from the original work, we enrich our dataset with annotations for depth~\cite{hu2024depthcrafter}, surface normals~\cite{DBLP:conf/eccv/KhirodkarBMZJSAS24}, and tracked human masks~\cite{ravi2024sam}, all generated using off-the-shelf predictors. This scalable pipeline yields a large-scale dataset ideal for training models to handle complex, multi-identity scenarios.

Extensive experiments demonstrate that our approach achieves state-of-the-art results in multi-identity human-centric video generation, preserving individual appearances and capturing realistic interactions. Our contributions are threefold: (1) A novel identity-embedding mechanism that ensures appearance consistency for multiple subjects by associating learnable tokens with segmentation masks. (2) A joint learning framework that models RGB, depth, and surface-normal maps in a shared latent space, enabling realistic 3D-aware human-object interactions. (3) The Multi-HumanVid dataset with rich annotations for  25K videos, designed to foster research in multi-person animation.

\section{Related Works}
\noindent\textbf{Human video generation.}
Human video generation seeks to produce consistent human videos starting from a single image. To improve controllability, most approaches in this domain leverage explicit human skeleton representations, \textit{e.g.}, OpenPose~\cite{DBLP:journals/pami/CaoHSWS21, simon2017hand, wei2016convolutional} and DensePose~\cite{DBLP:conf/cvpr/GulerNK18}, as supplementary guidance. Early methods predominantly relied on GANs for tasks such as image animation and pose transfer~\cite{chan2019everybody, ren2020deep, DBLP:conf/nips/SiarohinLT0S19, siarohin2019animating, siarohin2021motion, yu2023bidirectionally, zhao2022thin}. Recently, diffusion models~(DMs)~\cite{DBLP:conf/nips/HoJA20, DBLP:conf/iclr/SongME21, ni2023conditional, wang2023leo} have garnered attention in human image animation due to their impressive achievements and high-quality outputs in both image\cite{rombach2022high, nichol2021glide, ramesh2022hierarchical, DBLP:conf/nips/SahariaCSLWDGLA22, balaji2022ediffi, podell2023sdxl} and video~\cite{blattmann2023align,zhou2022magicvideo,singer2022make, ho2022video, ho2022imagen, ruan2023mm, yin2023dragnuwa, wang2023lavie, guo2023animatediff} generation. For example, MagicDance~\cite{chang2023magicdance} introduces a two-stage training approach that separates the learning of appearance from human motion. Animate Anyone~\cite{hu2023animate} employs a reference network to extract appearance features from the source image and integrates a motion module similar to AnimateDiff~\cite{guo2023animatediff} to maintain temporal consistency. Additionally, it includes a lightweight pose guider to encode pose information into the pre-trained models. In a similar vein, MagicAnimate~\cite{xu2023magicanimate} uses DensePose~\cite{DBLP:conf/cvpr/GulerNK18} for motion representation and incorporates ControlNet~\cite{zhang2023adding} to encode pose data. Champ~\cite{zhu2024champ} further enhances alignment by introducing the SMPL~\cite{DBLP:journals/tog/LoperM0PB15} model sequence along with rendered depth and normal maps. Human4DiT~\cite{shao2024human4dit} equips pose-driven human video generation with the ability of multi-view to perform 4D human video generation by training a Diffusion Transformer (DiT). CamAnimate~\cite{wang2024humanvid} incorporates camera pose control ability to the original human pose control and enables human video generation with simultaneous subject and camera movements in both training and inference. All previous mentioned methods focus on single-person pose-guided human video generation, and they overlook the generation of multiple subjects or interactions. To the best of our knowledge, our model is the first to generate human-centric interactions in videos.

\noindent\textbf{Human-centric Video Datasets.}
A diverse and extensive collection of human-centric video datasets is vital for advancing human image animation tasks. Among real-world datasets sourced from the Internet, TikTok~\cite{jafarian2021learning} offers 340 human-centric video clips from social media, featuring a wide array of appearances and performances, while UBC-Fashion~\cite{zablotskaia2019dwnet} comprises 500 fashion video clips set against plain backgrounds. Many synthetic human datasets also try to provide human images/videos and corresponding human poses and camera parameters by rendering engines, such as AGORA~\cite{DBLP:conf/cvpr/PatelHTHTB21}, HSPACE~\cite{bazavan2021hspace}, GTA-Human~\cite{cai2021playing}, SynBody~\cite{DBLP:conf/iccv/YangCML0XWQW0W023} and BEDLAM~\cite{bedlam}. Recently, the synthetic part of HumanVid~\cite{wang2024humanvid} renders individuals with physically realistic clothing appearances in 3D environments with diverse camera trajectories; while the internet part of HumanVid collects 20K high-quality human videos without interactions and leverages SLAM-based methods~\cite{wang2024tram} to fit camera trajectories. Building upon HumanVid, we further collect 25K human videos with multiple subjects and diverse interactions for generating human-centric interaction videos.
\begin{figure*}[t]
    \centering  
    \includegraphics[width=0.99\textwidth]{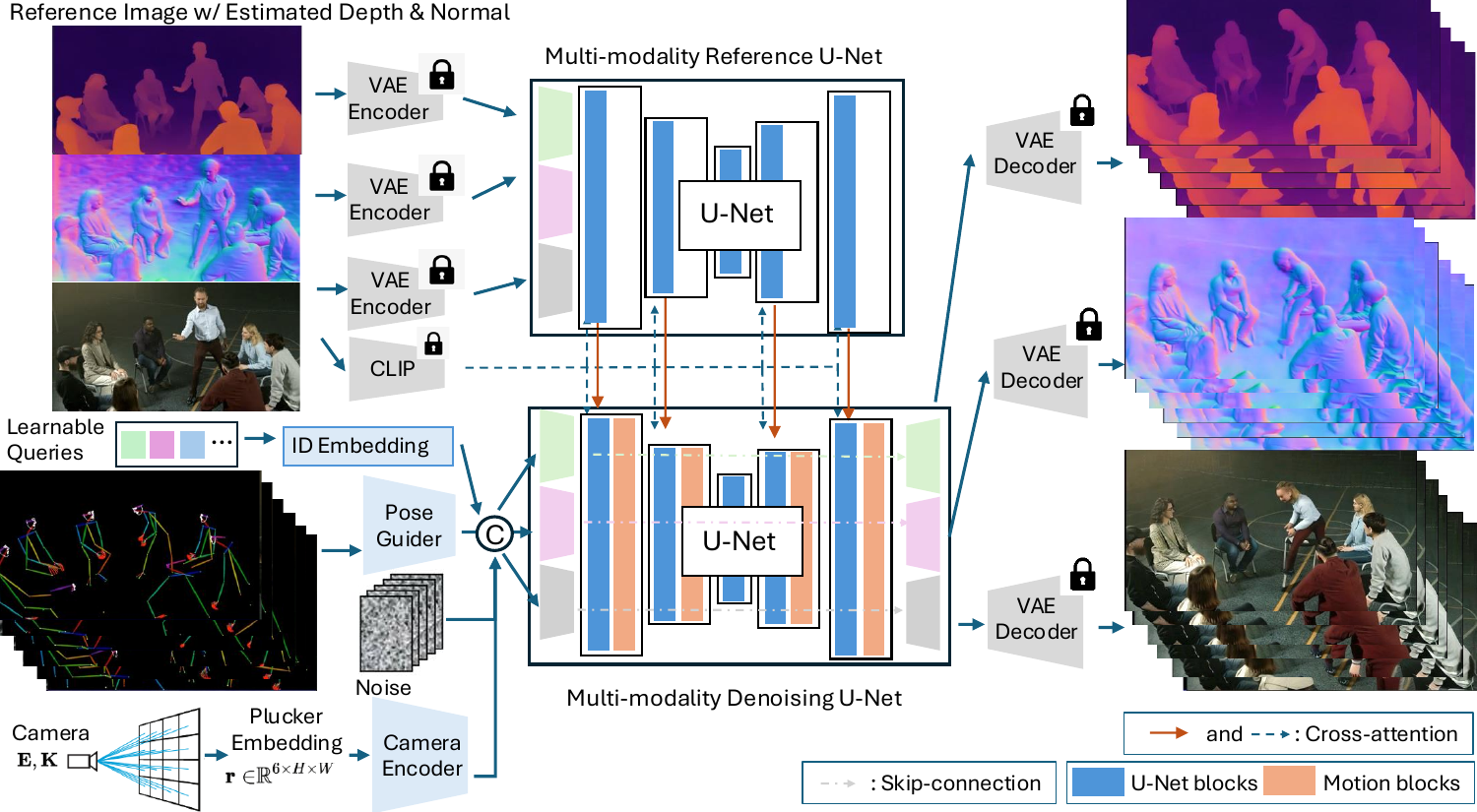}
    \caption{{\em Structural Video Diffusion} has two key components: Learnable ID embeddings and multi-modality structural information prediction (i.e., depth and surface-normal). Given ID embeddings from human masks, it is able to distinguish multiple people's pose conditions in interaction scenarios. It is also able to warp depth and normal information from the reference image to the entire video according to human pose and camera pose conditions.}
    \label{fig:framework}
    \vspace{-1em}
\end{figure*}

\section{Multi-identity Structural Video Diffusion}
Our goal is to generate human-centric videos with rich human-human and human-object interactions, from a reference image, and a sequence of driving camera parameters and 2D human poses with identities. The overall framework is shown in Fig.~\ref{fig:framework}.

\subsection{Preliminaries}
\label{sec:preliminaries}
\noindent\textbf{Video Diffusion Models.} Latent image-to-video diffusion models~\cite{rombach2022high,ho2022video,blattmann2023stable,guo2023animatediff} aim to learn the conditional distribution $p(\mathbf{x}|\mathbf{c})$ of encoded video data $\mathbf{x}$ (where $\mathbf{x} = \mathcal{E}(X)$, $X$ is the rgb video, and $\mathcal{E}(\cdot)$ denotes the VAE encoder~\cite{kingma2013auto}) conditioned on a image $\mathbf{c}$ in the latent space. The diffusion process applies a variance-preserving Markov chain~\cite{sohl2015deep,ho2020denoising,song2020score} to $\mathbf{x}_0$, gradually adding noise as described by:
g$\mathbf{x}_t = \sqrt{\Bar{\alpha}_t}\mathbf{x}_0 + \sqrt{1- \Bar{\alpha}_t}\boldsymbol{\epsilon}, \boldsymbol{\epsilon} \sim \mathcal{N}(\boldsymbol{0},\boldsymbol{I})$, for $t = 1, \dots, T$, with $T=1000$ and $\overline{\alpha}_t$ controlling noise levels. The denoising process predicts the noise $\boldsymbol{\epsilon}_{\theta}(\mathbf{x}_t,t,\boldsymbol{c})$ using a neural network (e.g., UNet~\cite{DBLP:conf/miccai/RonnebergerFB15} or Diffusion Transformer~\cite{peebles2023dit}) conditioned on image embeddings $\mathbf{c}$. It is trained to minimize the weighted mean squared error: $L = \mathbb{E}\left[\omega(t) \lVert \boldsymbol{\epsilon} - \boldsymbol{\epsilon}_{\theta}(\mathbf{x}_t, t, \mathbf{c}) \rVert_{2}^{2}\right]$, where $\omega(t)$ is a hyper-parameter that defines the weighting of the loss at timestep $t$. After training, it generates images by denoising from $\mathbf{x}_T \sim \mathcal{N}(\boldsymbol{0},\boldsymbol{I})$ to $\boldsymbol{\hat x}_0$ using a fast sampler~\cite{DBLP:conf/nips/0011ZB0L022, DBLP:conf/iclr/SongME21}, and decodes $\boldsymbol{\hat x}_0$ back to video $X$ with a VAE decoder $\mathcal{D}(\cdot)$.

\noindent \textbf{Task Formulation.} Given an input reference image embedding $\boldsymbol{c}=\mathcal{E}(C), C \in \mathbb{R}^{H \times W \times 3}$ of $N$ human identities $\{\mathbf{e}_n\}_{n=1}^N$ and their corresponding tracking masks $\mathbf{M}^f \in \{0,1,...,N\}^{H \times W}$, 2D human skeleton maps~\cite{DBLP:journals/pami/CaoHSWS21} $\mathbf{P}^f \in \mathbb{R}^{H \times W \times 3}$, and camera parameters $\mathbf{R}^f \in \mathbb{R}^{3 \times 4}$ for the $f$-th frame, our objective is to synthesize human-centric videos $X \in \mathbb{R}^{F \times H \times W \times 3}$ with multiple identities following the given human pose and camera pose condition. The overall conditional generation's formulation $f(\cdot)$ is
\vspace{-0.5em}
\begin{equation}
    f(\cdot): (\boldsymbol{c}, \{\mathbf{M}^f\}_{f=1}^F, \{\mathbf{P}^f\}_{f=1}^F, \{\mathbf{R}^f\}_{f=1}^F) \rightarrow \mathbf{x}.
\end{equation}
In practice, the diffusion process's output $\boldsymbol{\hat x}_0$ is used as prediction of actual video latent $\mathbf{x}$. A major difference with existing human image animation methods is that the reference image $\boldsymbol{c}$ and 2D human skeleton maps $\{\mathbf{P}^f\}_{f=1}^F$ could have multiple humans, and the association of a specific human appearance and 2D human pose is in the tracking masks $ \{\mathbf{M}^f\}_{f=1}^F$. 

\noindent \textbf{Camera Control.} We follow CameraCtrl~\cite{he2024cameractrl} and CamAnimate~\cite{wang2024humanvid} to adopt plucker embedding to represent camera parameters and then inject them to the Denoising Unet for controlling camera movement and human movement in the same time. As we build our method from CamAnimate~\cite{wang2024humanvid}, please refer to Sec. 3.3 in the original paper for more details about incorporating camera controls in video diffusion models.

\subsection{ID-Specific Human Image Animation}
\label{sec:id_specific_human_image_animation}
To address multi-identity video generation under the human-centric setting, we focus on preserving consistent appearances for $N$ distinct human identities throughout dynamic scenes, where individuals move and exchange positions. From task formulation in Sec.~\ref{sec:preliminaries}, each tracking mask $\mathbf{M}^f$ encodes the identity labels of pixels across $N$ individuals at frame $f$. Our key objective is to incorporate identity-specific features into the model so that person $n$ in the reference image $\boldsymbol{c}$ is consistently mapped to the same person $n$ in the generated frames, even if positions change over time. To achieve this, we propose learnable identity embeddings to effectively associate them.

\noindent \textbf{ID-Embedding via Human Masks.}
Inspired by detection transformers~\cite{carion2020end}, we introduce a set of $N$ learnable ID embeddings $\mathbf{E}_{query} \in \mathbb{R}^{N \times C}$. These embeddings serve as identity tokens for the $N$ individuals. Specifically, for each frame $f$, we convert $\mathbf{M}^f$ into a spatial ID-embedding map, $\mathbf{E}^f \in \mathbb{R}^{H \times W \times C}$, by copying the $n$-th embedding row of $\mathbf{E}_{query}$ to all spatial locations $(h,w)$ in $\mathbf{E}^f$ where $\mathbf{M}^f(h,w) = n$. Consequently, the resulting map $\mathbf{E}^f$ has the same spatial size as each video latent feature, but each human pixel location is filled with the corresponding identity embedding. This procedure binds each human’s appearance identity (from the reference image $\boldsymbol{c}$) to its spatial region in the latent space. For a video, the final ID-embedding map is $\mathbf{E} \in \mathbb{R}^{F \times H \times W \times C}$.

\noindent \textbf{Integration with the Denoising Process.}
We integrate the ID-embedding map $\mathbf{E}$ with the noisy latent $\mathbf{x}_t$ in a ControlNet manner~\cite{zhang2023adding} inside the denoising network at each diffusion step $t$: $\widetilde{\mathbf{x}}_t = \mathbf{x}_t + {\text zero\_conv}(\mathbf{E})$, where $\widetilde{\mathbf{x}}_t$ is the updated noisy latent and $zero\_conv$ is a zero-initialized convolution~\cite{zhang2023adding}. This operation ensures that the network is identical to a single-person human image animation method at the beginning of multi-human training. Such identity-specific clues injected by $\mathbf{E}$ at each spatial location could further guide the model learns to consistently preserve person $n$’s appearance across frames by conditioning on these ID embeddings. 

The ID-specific embedding mechanism allows for up to $N$ unique identities, each tracked through $\mathbf{M}^f$. If fewer subjects appear in the scene, the unused embeddings are simply ignored. This design offers a flexible means to handle diverse numbers of individuals in a single temporal framework. Crucially, it requires only one reference image embedding $\boldsymbol{c}$ for all identities while permitting extensive spatiotemporal transformations in the final video, thus enabling robust, multi-identity video generation with coherent human appearances.

\subsection{Latent Structural Video Diffusion}
\label{sec:latent_structural_video_diffusion}
To tackle the challenge of modeling human-object interactions in a video diffusion framework, we propose to jointly synthesize RGB, depth, and surface-normal representations, i.e., $\{\mathbf{x}_\mathrm{rgb}, \mathbf{x}_{\mathrm{depth}}, \mathbf{x}_{\mathrm{normal}}\}$ and treat their joint distribution as the distribution of prediction target $\mathbf{x}$. Leveraging such additional geometric maps effectively captures the underlying 3D structure of both humans and surrounding objects, enabling coherent movement and interaction in complex scenarios. Unlike methods that rely on frame-wise depth or normal annotations \emph{as inputs} (which are rarely available in real-world applications), we treat them as output modalities alongside the RGB domain. During training, off-the-shelf depth-\cite{hu2024depthcrafter} and normal-estimation~\cite{DBLP:conf/eccv/KhirodkarBMZJSAS24} tools automatically generate structural labels for each video frame, and the model learns to predict these geometric representations from the given human poses. As a result, it can more robustly infer object location changes with human motions without requiring object annotations.

\noindent\textbf{Structural Multi-modality Branches with Shared Backbone.}
The diffusion UNet's architecture consists of three main components: down-sampling blocks, middle blocks, and up-sampling blocks, with convolution and self-/cross-attention layers placed between them. The \textit{DownBlock}s compress noisy input data into lower-resolution hidden states, while the \textit{UpBlock}s expand these features to predict noise. Inspired by the image-based multi-branch denoising framework~\cite{hyperhuman}, we adopt a unified diffusion backbone while introducing dedicated \emph{expert branches} for each modality (RGB, depth, normal). Specifically,

\noindent\emph{(1) Multi-modality Denoising UNet:} We replicate the \textit{conv\_in}, \textit{conv\_out}, the first layer of \textit{DownBlock}s, and the last layer of \textit{UpBlock}s for each branch, so the modality-specific inputs (noisy $\mathbf{x}^t_{\mathrm{rgb}}$, $\mathbf{x}^t_{\mathrm{depth}}$, $\mathbf{x}^t_{\mathrm{normal}}$) and outputs (predicted noise in each modality) remain spatially aligned. Layers in the middle are shared to ensure a joint representation, and the model can capture cross-modality correlations.

\noindent\emph{(2) Multi-modality Reference UNet:} In addition to the main denoising UNet, we utilize a reference UNet to handle the identity clues and frame-wise pose and camera embeddings (Sec.~\ref{sec:id_specific_human_image_animation}). We likewise replicate the initial downsampling layers (i.e., \textit{conv\_in} and the first layer of \textit{DownBlock}s) for RGB, depth, and normal extracted from the reference image, combined with ID-specific embeddings. By introducing multi-modal cues in the reference image in the Reference UNet, the denoising UNet could warps depth and normal from the reference image rather than \emph{predicting} them, alleviating its burden to simultaneously infer geometry, especially when we finetune the base model to multi-modal prediction in a human-centric video dataset that is much smaller scale than the pretraining video dataset.

Such design allows each modality to have its own enter/exit pathways in the network, ensuring the final video outputs remain spatially consistent across RGB, depth, and normal channels. Meanwhile, the shared backbone layers help unify shape details and geometric cues, guiding how objects move with or remain static relative to human poses depending on the spatial relationship with human poses.

\noindent\textbf{Learning Objective.}
We train our video diffusion model to predict the noise for all three modalities together. Similar to~\cite{ho2020denoising}, we sample independent Gaussian noise $\boldsymbol{\epsilon}_{\mathbf{x}_{\mathrm{rgb}}}, \boldsymbol{\epsilon}_{\mathbf{x}_{\mathrm{depth}}}, \boldsymbol{\epsilon}_{\mathbf{x}_{\mathrm{normal}}}\sim \mathcal{N}(\mathbf{0},\mathbf{I})$ and diffuse each modality with a variance-preserving schedule. Let $t\in \{1,\dots,T\}$ be the diffusion timestep, and noise $\widehat{\boldsymbol{\epsilon}}_{\boldsymbol{\theta}}(\cdot)$ be our unified network’s prediction. As timestep $t$ is identical for three modalities in the inference process, we also sample the same $t$ for them in the training process. In practice, we utilize $\mathbf{v}$-prediction as the training target, i.e., ${\mathbf{v}}^t_{\mathbf{x}_{m}}=\alpha_t \boldsymbol{{\epsilon}}_{\mathbf{x}_{m}}-\sigma_t \mathbf{x}_{m}$, for $m \in \{\text{rgb, depth, normal}\}$. The final training objective then becomes:
\begin{equation}
\label{eq:ldepthnormalrgb}
\small
\begin{aligned}
\mathcal{L}(\boldsymbol{\theta}) \;=\; 
\mathbb{E}_{
\substack{
\mathbf{x},
\boldsymbol{c}, \{\mathbf{M}^f,\mathbf{P}^f,\mathbf{R}^f\}_{f=1}^F,
\mathbf{v}_{\mathbf{x}}, t
}} \biggl[
        \;\bigl\|\mathbf{v}_{\mathbf{x}^t_{\mathrm{rgb}}} \;-\; \widehat{\mathbf{v}}_{\boldsymbol{\theta}, \text{rgb}} 
    \bigr\|_2^2 \\
    +\;\bigl\|\mathbf{v}_{\mathbf{x}^t_{\mathrm{depth}}} \;-\; \widehat{\mathbf{v}}_{\boldsymbol{\theta}, \text{depth}} 
    \bigr\|_2^2 
    +\;\bigl\|\mathbf{v}_{\mathbf{x}^t_{\mathrm{normal}}} \;-\; \widehat{\mathbf{v}}_{\boldsymbol{\theta}, \text{normal}} 
    \bigr\|_2^2
\biggr],
\end{aligned}
\end{equation}
where $\widehat{\mathbf{v}}_{\boldsymbol{\theta}, \text{modal}} = \widehat{\mathbf{v}}_{\boldsymbol{\theta}}\!\bigl(
        \mathbf{x}^t_{\mathrm{modal}},\;\boldsymbol{c},\;\{\mathbf{M}^f,\mathbf{P}^f,\mathbf{R}^f\},
        \;t \bigr)$ for modal in \{rgb, depth, normal\}, 
By jointly denoising all modalities, the model learns a single coherent representation that captures the geometry of both humans and objects (via $\mathbf{x}_{\mathrm{depth}}$ and $\mathbf{x}_{\mathrm{normal}}$) and the appearance details (via $\mathbf{x}_{\mathrm{rgb}}$), thus substantially improving human-object interaction quality in the synthesized videos. It is worth noting that our approach \emph{does not} require explicit object-level conditions. Instead, by leveraging depth and normal predictions for every video frame, the model infers object positions, orientations, and motion based on how humans interact with them. Even though this remains limited compared to complete physical simulations, it represents a pioneer step toward more realistic human-centric video generation in complex 3D scenarios.

\section{Data Preparation}
\label{sec:data}
\noindent {\bf Data Curation from Pose Estimation.} Following HumanVid~\cite{wang2024humanvid}, we collected data by querying the \textsc{Pexels} API~\cite{Pexels} with interaction-centric keywords such as party, and curate videos using 2D human pose detection~\cite{yang2023effective}. During the data curation process, we focused on several key metrics: average confidence scores for upper body keypoints $c$, the ratio $r$ of frame space occupied by the largest detected person, and the average number of people per frame ($n$). We adopt the following criteria: the human should be clear in the video ($c > 0.5$), the primary subject must occupy a significant portion of the frame ($r > 0.07$), and the scene should not be crowded ($n \leq 5$). We collect 25K more human-centric videos for training, extending the total training data size of our method to be 45K videos, while the existing HumanVid~\cite{wang2024humanvid} dataset only has 20K videos. 

\noindent{\bf Human Mask Tracking.} We utilize Grounding-DINO~\cite{liu2025grounding} with word `human' as query to ground human bounding boxes in keyframes and leverage SAM2~\cite{ravi2024sam} to track such human masks by taking human bounding boxes as input. After tracking the entire video, a post-processing will be adopted to reverse track and merge mask identities. Thanks to the grounding model operated on keyframes, we could also track humans that appears later in the videos. 

\noindent{\bf Camera Trajectory Estimation.} Following HumanVid~\cite{wang2024humanvid}, we adopt TRAM~\cite{wang2024tram} to utilize a SLAM method~\cite{DBLP:conf/nips/TeedD21} for recovering camera extrinsic parameters from in-the-wild videos with explicit human movement. To ensure camera parameters are robust to dynamic humans, we employ the human masks estimated in the above step to remove dynamic regions in camera estimation. As the model is agnostic to camera intrinsics, we configure the SLAM system to estimate several pre-defined camera intrinsics to find the best one according to SLAM errors. To produce metric-scale camera estimations, we leverage semantic cues by utilizing noisy depth predictions~\cite{bhat2023zoedepth}. 

\noindent{\bf Video Depth Estimation.} We utilize Depthcrafter~\cite{hu2024depthcrafter} to extract non-metric video-level depth maps and visualize them with color maps. Depthcrafter is finetuned from a video diffusion model~\cite{blattmann2023stable}, therefore the temporal consistency of depth prediction is greatly improved over image-based depth estimation methods.

\noindent{\bf Surface-normal Estimation.} We utilize Sapiens~\cite{DBLP:conf/eccv/KhirodkarBMZJSAS24} to extract surface-normal maps in the human regions via human masks obtained in the above step. As Sapiens is only pretrained on human-centric data, it predicts worse normal maps for background regions, so we only utilize the surface-normal maps within human masks in our method. 
\section{Experiments}
\label{sec:experiments}
We utilize the evaluation protocol of HumanVid~\cite{wang2024humanvid} on our collected interaction-centric human videos, i.e., videos with human-human or human-object interactions. Our test set has 80 human-centric videos in total. Due to that interactions are inherently complex in the temporal dimension, so we evaluate videos in a longer temporal interval. For all models compared in this section, we predict frames in the range [1,144) with a stride of 3, resulting in a sequence of 48 frames. We use the middle frame of a sequence as the reference image. We evaluate each video under this setting using PSNR~\cite{DBLP:journals/tip/WangBSS04}, SSIM~\cite{DBLP:journals/tip/WangBSS04}, LPIPS~\cite{DBLP:conf/cvpr/ZhangIESW18}, FID~\cite{DBLP:conf/nips/HeuselRUNH17}, and FVD~\cite{DBLP:conf/iclr/UnterthinerSKMM19} metrics. We use the Internet data part of HumanVid~\cite{wang2024humanvid} and our collected Multi-HumanVid for training.

\begin{figure*}[t]
    \centering  
    \includegraphics[width=0.99\textwidth]{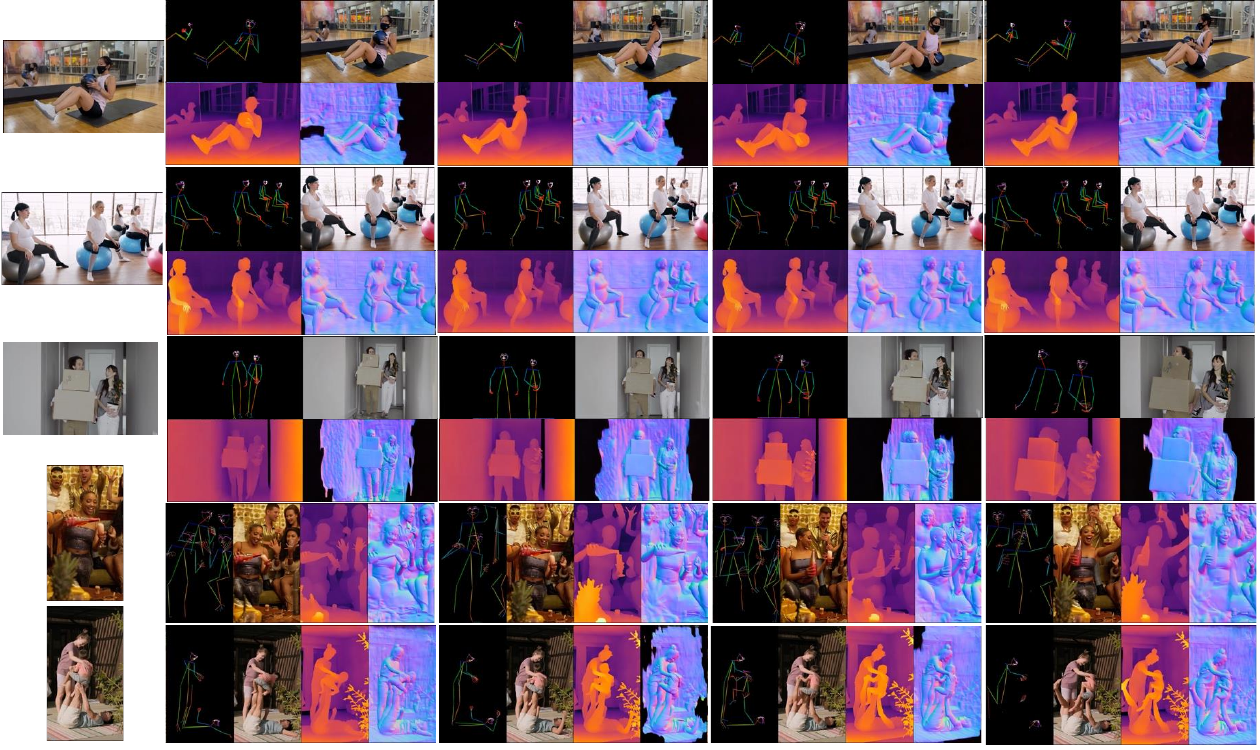}
    \caption{Qualitative results of our Structural Video Diffusion, which could generate complex human-human or human-object interactions. The first image is the reference image, and each human pose image in the sequence is the input condition, while other three images in the sequence is our results. It is worth noting that the normal supervision is only adopted within human regions, so the discontinue normal map predictions do not affect our RGB predictions. }
    \vspace{-1em}
    \label{fig:qualitative}  
\end{figure*}

\noindent{\bf Implementation Details.} We initialize the Denoising UNet and ReferenceNet with the Stable Diffusion 1.5 checkpoint~\cite{rombach2022high}, and the Pose Guider with ControlNet~\cite{zhang2023adding} weights trained on OpenPose~\cite{DBLP:journals/pami/CaoHSWS21}. The camera encoder is initialized using weights from CameraCtrl~\cite{he2024cameractrl}. We train on a mixture of horizontal and vertical videos with resolutions $(896, 512)$ and $(512, 896)$, respectively. Each batch contains only horizontal or vertical videos, selected randomly between batches, to balance visual quality and computational cost based on GPU memory constraints.
In the first stage, all network parameters are trained with a batch size of $8$ (without depth and normal), $5$ (with one additional modality), or $4$ (with both depth and normal), depending on GPU memory limitations. In the second stage, we freeze the Denoising UNet, ReferenceNet, and Pose Guider, and train only the camera encoder and motion module. The motion module is initialized with AnimateDiff~\cite{guo2023animatediff} V3 weights. The second stage uses a batch size of $1$ and processes $24$ frames (without depth and normal), $21$ frames (with one additional modality), or $16$ frames (with both depth and normal).
Training is conducted on $8$ NVIDIA A100 GPUs for all stages and $1$ NVIDIA A100 GPU for testing. The first and second stages are trained for $40{,}000$ and $20{,}000$ iterations, respectively, using a learning rate of $1\text{e-}5$ and the AdamW optimizer~\cite{loshchilov2017decoupled}. Our camera embedding uses the Plücker embedding from CameraCtrl, derived from the camera's intrinsic and extrinsic parameters.

\begin{table}[t]
\centering
\caption{Comparison with SOTA on our Multi-ID test set.}
\vspace{-0.5em}
\resizebox{0.49\textwidth}{!}{%
\begin{tabular}{l|cccccc}
\toprule
\textbf{Methods} & SSIM $\uparrow$ & PSNR $\uparrow$ & LPIPS $\downarrow$ & FVD $\downarrow$ & FID $\downarrow$ \\ 
\midrule
MimicMotion~\cite{zhang2024mimicmotion}  & 0.628 & 19.878 & 0.258 & 1042.6 & 59.11\\
CamAnimate~\cite{wang2024humanvid}   & 0.649 & 19.552 & 0.265 & 982.1 & 54.09\\
Ours  &{\bf 0.691} &  {\bf 20.685} &  {\bf 0.233} &  {\bf 878.2}  &  {\bf 30.57}\\ 
\bottomrule
\end{tabular}
}
\label{tab:sota}
\vspace{-0.5em}
\end{table}

\subsection{Human-centric Interaction Generation}
\noindent{\bf Quantitative comparison with previous methods.} In Tab.~\ref{tab:sota}, we compare our method with two previous state-of-the-art method MimicMotion~\cite{zhang2024mimicmotion} and CamAnimate~\cite{wang2024humanvid} on our collected multi-human videos with rich human-human and human-object interactions. Due to that MimicMotion cannot generate videos that precisely follow the camera movement, it achieves the worst performance on most of the metrics. As CamAnimate is not designed specifically for generating videos of multiple identities, its performance is much worse than ours in such hard scenarios with interactions, although our method shares the same video generation foundation model~\cite{guo2023animatediff} with it. This result indicates that our design of ID-embedding and structural video diffusion is effective in modeling complex interaction scenarios. Such pseudo-3D information helps the model to better learn the spatial locations of humans and objects when multiple identities are interacted with each other.

\noindent{\bf User-study.} Since our model also accounts for camera movements, we perform a qualitative comparison with CamAnimate~\cite{wang2024humanvid} using a questionnaire consisting of 10 single-choice questions. A total of 20 participants took part in our user study, and our method received a dominant preference of \textbf{91.25\%} over the competing approach. Please refer to our supplementary materials for more videos used in our user-study.

\noindent{\bf Qualitative results.} In Fig.~\ref{fig:qualitative}, we show qualitative results of our structural video diffusion in joint generation of RGB, depth and normal maps. By leveraging such ability, we show that our model could animate human videos with multiple identities and complex interactions with objects and others. In our model, we could correctly associate human appearances with driving poses and also maintain the appearance of objects during its motion process. Please refer to our supplementary materials for more video results.

\begin{table}[t]
\centering
\caption{Ablation study on key components of our method.}
\vspace{-0.5em}
\resizebox{0.49\textwidth}{!}{%
\begin{tabular}{l|cccccc}
\toprule
\textbf{Methods} & SSIM $\uparrow$ & PSNR $\uparrow$ & LPIPS $\downarrow$ & FVD $\downarrow$ & FID $\downarrow$ \\ 
\midrule
Baseline~\cite{wang2024humanvid}   & 0.649 & 19.552 & 0.265 & 982.1 & 54.09\\ 
+ ID-embedding    & 0.686 & 20.374 & 0.237 & {\bf 873.5} & 33.75\\ 
+ Multi-modality  & 0.668 & 20.139 & 0.240 & 907.8 & 47.67\\ 
+ Both & {\bf 0.691} &  {\bf 20.685} &  {\bf 0.233} &  878.2  &  {\bf 30.57}\\ 
\bottomrule
\end{tabular}
}

\label{tab:compoents}
\vspace{-0.5em}
\end{table}

\subsection{Cross-Identity Motion Transfer}
\begin{figure*}[t]
    \centering  
    \includegraphics[width=0.97\textwidth]{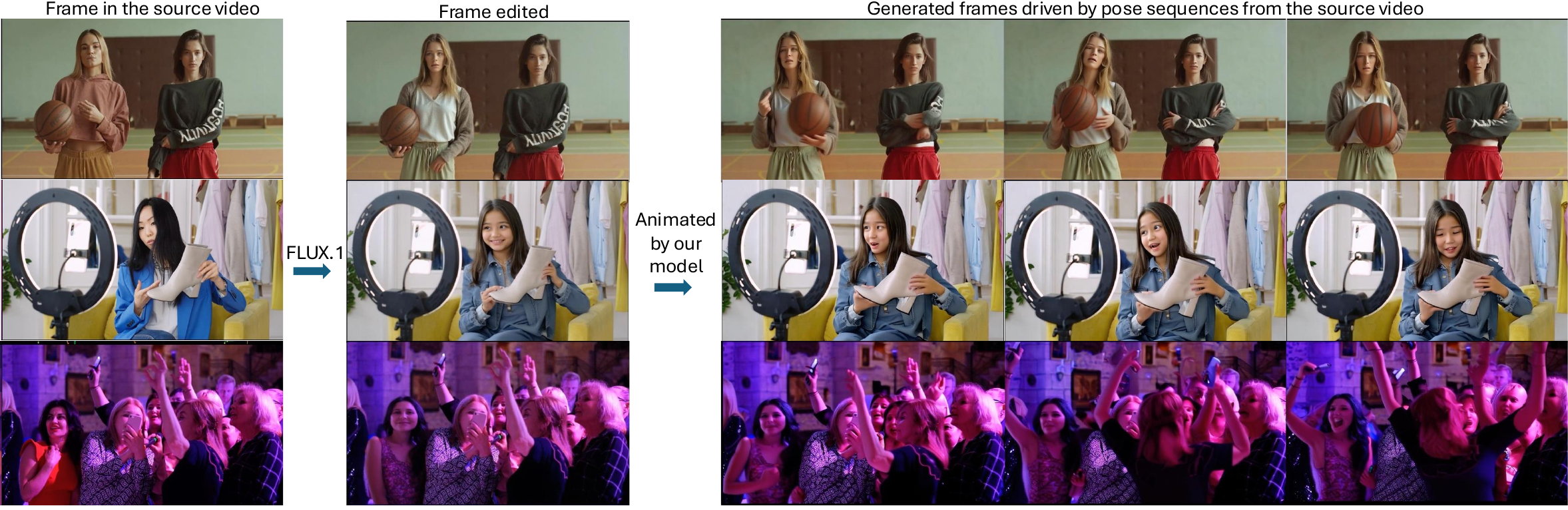}
    \caption{Qualitative results of cross-identity human image animation. One of the person's appearance is edited by FLUX.1~\cite{flux2024}. }
    \label{fig:cross-id}
    \vspace{-1em}
\end{figure*}

As animating the human-centric videos from a single human image is the major focus of this area, we show our cross-identity results by adopting 2D human pose sequences from a source video and leverage a reference image containing human appearance edited by image inpainting methods such as FLUX.1~\cite{flux2024}. By animate edited human images, we could effective transfer the original motion templates to novel appearances, as shown in Fig.~\ref{fig:cross-id}.

\subsection{Ablation Study}
\begin{table}[t]
\centering
\caption{Ablation study on predicted modalities.}
\vspace{-0.5em}
\resizebox{0.49\textwidth}{!}{%
\begin{tabular}{l|cccccc}
\toprule
\textbf{Methods} & SSIM $\uparrow$ & PSNR $\uparrow$ & LPIPS $\downarrow$ & FVD $\downarrow$ & FID $\downarrow$ \\ 
\midrule
RGB-only   &  0.686 & 20.374 & 0.237 & {\bf 873.5} & 33.75\\ 
+ Depth  &{\bf 0.691} &  {\bf 20.685} &  {\bf 0.233} &  878.2  &  {\bf 30.57}\\ 
+ Normal  & 0.639 & 19.037 & 0.272 & 924.8 & 60.58\\ 
+ Depth\&Normal & 0.643 &  19.664 &  0.264 &  898.7  &  56.78\\ 
\bottomrule
\end{tabular}
}

\label{tab:modality}
\vspace{-0.5em}
\end{table}

\begin{figure}[h]
    \centering
    \includegraphics[width=0.49\textwidth]{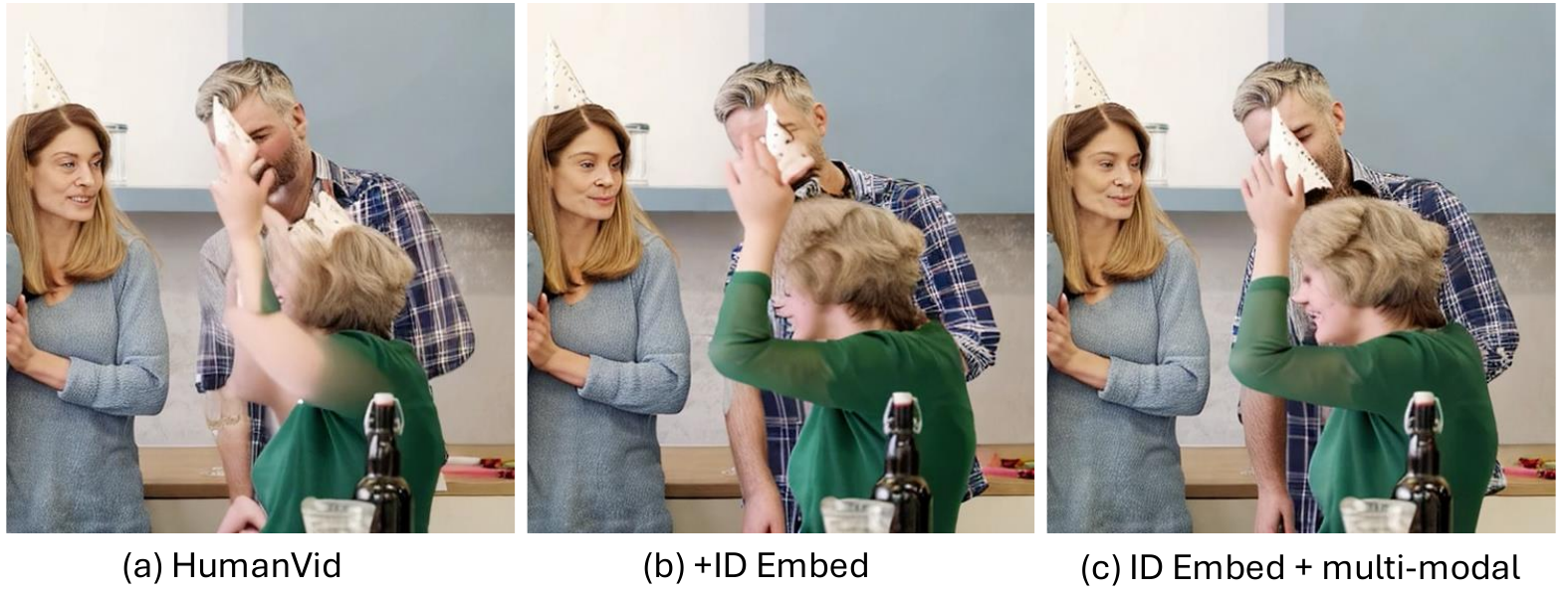}
    \caption{Qualitative ablation of ID embedding and depth/normal modality. ID embedding improves inaccurate or lost pose keypoints and depth/normal enhances the 3D consistency. }
    \label{fig:qualitative-compare}
    \vspace{-1em}
\end{figure}

\noindent{\bf Ablation on ID-embedding and Structural Learning.} In Tab.~\ref{tab:compoents}, we ablate the contribution of two components proposed in our paper: ID-embedding and Structural Learning. Due to the test set contains many human-human interactions, ID-embedding itself could already lead to a slightly better performance. Similarly, letting the model to learn the joint distribution of RGB and pseudo-3D information could also enhance the quality of human video generation, especially when people are moving fast or interacting with objects. By combining two components together, our method achieves a notable performance gain over the strong baseline method~\cite{wang2024humanvid}. Experiments show that previous single-person human image animation methods could not properly generate realistic multi-identity human interactions, and our method shows a potential approach for better interaction generation in human videos.

\noindent{\bf Ablation on Modalities in Structural Learning.} In Tab.~\ref{tab:modality}, we ablate the importance of each modality in our structural video diffusion framework. We find that generally depth contributes more to the final performance than the surface-normal. As the normal annotations are only effective in human regions, we only utilize them within estimated human masks as supervision. This method cannot learn complete distribution of surface normal maps over the entire videos, thus limits its ability to better generate multi-human videos. Due to the low quality of normal estimation method~\cite{DBLP:conf/eccv/KhirodkarBMZJSAS24} in our dataset preparation process, the normal maps could even make the video generation process more noisy and produce worse performance according to Tab.~\ref{tab:modality}. On the contrary, the depth estimation method~\cite{hu2024depthcrafter} in our dataset preparation is finetuned from video generation model~\cite{blattmann2023stable}, therefore it could predict smooth depth information for the entire video. It provides more complete pseudo-3D information for our human image animation model to learn the 3D-aware appearance distribution about human interactions in videos. We utilize our model with only depth annotation as default. However, we still believe surface-normal maps could contribute to video generation task if normal annotation methods in the future could be more accurate and stable in the temporal dimension.
\section{Conclusion}
In this paper, we address the challenge of generating multi-identity human-centric videos from a single reference image by introducing {\em Structural Video Diffusion}. Our method incorporates two core ideas for modeling human-human and human-object interactions. First, we design a learnable ID-embedding scheme that assigns separate embeddings to different individuals, thereby preserving consistent appearances throughout videos even when subjects change positions or overlap in the camera frame. Second, we incorporate pseudo-3D structural information (i.e., depth and surface-normal maps), into a multi-modality diffusion network, enabling the model to capture and animate intricate human-object interactions. We expand the existing human-centric video dataset with 25K additional videos containing rich multi-identity scenes and diverse pose interactions. Through comprehensive experiments, our approach demonstrates superior fidelity, realism, and temporal consistency in generating human-centered videos with multiple subjects and objects, outperforming various single-human baselines. This work provides a pioneer attempt for video generation with complex identities and interactions, and we hope it could drive further progress in realistic content creation.

\noindent{\bf Limitations.} Due to the limited computational resources, we cannot implement our idea in large video diffusion transformers such as HunyuanVideo~\cite{kong2024hunyuanvideo} or CogVideoX~\cite{yang2024cogvideox}, leading to suboptimal visual qualities and unstable pixel motions in the video generation results. 

\noindent{\bf Acknowledgment.} This project is funded in part by Shanghai Artificial Intelligence Laboratory, CUHK Interdisciplinary AI Research Institute, the Centre for Perceptual and Interactive Intelligence (CPII) Ltd under the Innovation and Technology Commission (ITC)’s InnoHK, and HKU Startup Fund.
{
    \small
    \bibliographystyle{ieeenat_fullname}
    \bibliography{main}
}

\end{document}